\documentclass{article}

\usepackage[final]{corl_2026} 
\usepackage{graphicx}
\usepackage{wrapfig}
\usepackage{float}
\usepackage{amsmath}
\usepackage{amssymb}
\usepackage{newtxmath}
\usepackage{booktabs}
\usepackage[font=small]{caption}
\usepackage{makecell}
\hypersetup{pdftitle={Learning to Be Different: Component Specialization for Compositional Diffusion Policies},
            pdfauthor={Alec Laprevotte, Haonan Chen, Xiaoshen Han, Yilun Du}}
\usepackage{tikz}
\usetikzlibrary{arrows.meta}
\usepackage{xcolor}
\newsavebox{\liberotablebox}
\newsavebox{\liberoplotbox}
\newlength{\liberofigwidth}

\title{Learning to Be Different: Component Specialization for Compositional Diffusion Policies}

\author{
  Alec Laprevotte, Haonan Chen, Xiaoshen Han, Yilun Du\\
  Harvard University 
}

\begin{document}
\maketitle

\vspace{-1.5em}
\begin{abstract}
    Diffusion policies with linearly composed scores have emerged as a promising approach to multitask robotic learning by combining separate factors to enable modular, task-wise adaptation, compositional generalization, and interpretable functional decomposition. However, current training objectives do not encourage factor specialization, often resulting in uninterpretable denoisers whose composed prediction is an average over identical copies, undermining their architectural premise. To address this, we propose a drop-in score-decorrelation auxiliary loss, motivated by projective composition theory. Our approach encourages specialization by penalizing overlap directly between component noise predictions, with little architectural change or computational overhead. We demonstrate across multiple  benchmarks that our decorrelation loss consistently reduces measurable double-counting between components, improves task decomposition, and produces behaviorally meaningful specialization without compromising multitask performance. Website: \url{https://al3cl1.github.io/robotics-score-decorrelation}.
\end{abstract}

\keywords{Diffusion Policies, Multitask Learning, Robot Manipulation} 


\section{Introduction}
Offline imitation learning has become one of the dominant paradigms to replicate complex robotic manipulation skills~\citep{chi2024diffusion, team2024octo, kim2024openvla}. However, as the diversity of tasks scales, their action distributions become highly multimodal. Monolithic policies model these behaviors jointly, merging them together and complicating both specialized skill interpretation and recombination across tasks~\citep{zitkovich2023rt, kim2024openvla, chen2025bimanual}.

To address these limitations, a growing line of work instead breaks down larger monolithic policies into smaller, composable generative models for better interpretability and test-time generalization~\citep{du2024compositional}. Particularly, many robotics models sum multiple diffusion components via linear score combination, breaking down complex action distributions into simpler, specialized factors that are aggregated at inference time~\citep{wang2024poco, reuss2025efficient, wang2024sparse, chen2025multi, liu2026flexible, huang2025moe}. While these methods provide architectural factorization, their training objective supervises only the composed sum of component scores, leaving the decomposition itself weakly constrained. As a result, components often collapse into overlapping predictors~\citep{liu2026flexible, wang2024sparse}, undermining the modular adaptation, compositional generalization, and interpretability that motivated factorization in the first place.

We propose an auxiliary loss that directly encourages the components of a compositional diffusion policy to specialize. During training, it penalizes overlap between the noise predictions of each pair of components, pushing them toward distinct contributions to the composed score. The loss is a drop-in addition to the standard denoising objective: it requires no change to the component networks and is discarded at inference. Our approach is motivated by recent theory on projective composition of diffusion models~\citep{bradley2025mechanisms}, in which Factorized Conditionals, a sufficient condition for projective composition, imply orthogonality between background-relative score deltas. Because compositional policies rarely include an explicit background model, we use output decorrelation as a practical proxy for this structure. Unlike orthogonality regularizers for mixture-of-experts models, which act on hidden activations~\citep{guo2026advancing, hendawy2024multi}, our loss acts directly on the score outputs, where overlap between components has a direct generative meaning.

We validate our approach on Factorized Diffusion Policy (FDP)~\citep{liu2026flexible} across RLBench~\citep{james2020rlbench}, RoboCasa~\citep{robocasa365paper}, and LIBERO~\citep{liu2023libero}, and further demonstrate its benefits on a real-world cup-flipping task. Our contributions are summarized as follows: (1) We introduce a training-time decorrelation loss that encourages specialization among the components of compositional diffusion policies, motivated by projective composition theory. (2) We show that it consistently reduces redundancy between components across factor counts and outperforms alternative decorrelation objectives, while maintaining or improving task success, especially on diverse and multi-step tasks. (3) We show that the resulting components specialize to behaviorally meaningful parts of the action space, such as the mobile base and the gripper, without supervision over the action space.

\section{Related Works}
\textbf{Compositional Diffusion Policies in Robotics:} Diffusion models have driven major progress in image and video generation~\citep{ho2020denoising, rombach2022high}, and have been rapidly adopted in robotics as flexible policy representations.   Diffusion Policy~\citep{chi2024diffusion} models a visuomotor policy as a conditional denoising process over action chunks and has since been extended with richer input representations~\citep{ze20243d, wang2024equivariant}, score-based goal conditioning~\citep{reuss2023goal}, and trajectory-level planning~\citep{janner2022diffuser, ajay2023conditional, chen2024diffusion}. More recently, scaling these policies to multitask regimes has motivated architectures designed around reuse and specialization of sub-modules to better aid in generalization. One line of work introduces a mixture-of-experts (MoE) structure with sparse routing over noise levels or skills~\citep{wang2024sparse, reuss2025efficient, huang2025moe, hao2026abstracting}. However, these approaches inherit well-documented pathologies of MoE training: an auxiliary load-balancing loss, usually meant to prevent expert collapse, also drives overlapping gradient flow across experts, homogenizing their representations and suppressing specialization~\citep{guo2026advancing, dai2024deepseekmoe, fedus2022switch}, while sparse routing introduces training instability and load imbalance~\citep{ fedus2022switch}, and hard expert selection makes individual experts difficult to reuse across different, unseen tasks. In response, a complementary line of models addresses these limitations by composing diffusion policies via linear score combination~\citep{liu2022compositional, du2023reduce}. Components can be separately trained and aggregated at inference~\citep{wang2024poco, cao2025modality, cao2025compose}, or jointly trained with learned routers (or uniform weights) over per-component noise predictions~\citep{liu2026flexible, chen2025multi}. Trajectory-level compositional diffusion follows a similar template, summing per constraint or per-skill scores to recombine behaviors at planning time~\citep{luo2026generative, yang2023compositional, urain2023se, ajay2023compositional}.

However, a common pattern across this latter lineage is that compositionality is enforced architecturally, but only weakly enforced functionally. The composed score is supervised end-to-end, so the decomposition itself is weakly identified: components can collapse to redundant predictors of the same conditional, which limits the generalization that composition is meant to provide. Our work addresses this gap by regularizing component outputs during training, motivated by \citet{bradley2025mechanisms}, whose Factorized Conditionals, a sufficient condition for projective composition, imply orthogonal background-relative score deltas.

\textbf{Expert Specialization and Decorrelation:} Decorrelation and orthogonality losses are well studied across several model families, each operating on a different representational substrate. Self-supervised learning penalizes the cross-correlation of hidden activations to prevent feature collapse~\citep{zbontar2021barlow, bardes2021vicreg, cogswell2015reducing}, while classical disentanglement penalizes statistical dependence among latent codes~\citep{higgins2017betavae, chen2018isolating, eastwood2018framework}. In MoE models, orthogonality is often applied to intermediate FFN outputs of co-activated experts, router weights, or learned routing distributions~\citep{ omi2025load, xiao2025instructmole, dai2024deepseekmoe, su2026variational}. Parameter-level approaches enforce orthogonality between LoRA residuals or task vectors~\citep{ilharco2022editing, po2024orthogonal, ortiz2023task}. In diffusion specifically, prior work enforces or analyzes near-orthogonal concept embeddings or representation subspaces~\citep{wang2023concept, geng2024factorized}. In robotics, classical compositional policies multiply or blend primitives without enforcing decorrelation~\citep{peng2019mcp, yang2020multi}, while, outside robotics, multi-branch architectures orthogonalize at the feature level for adversarial robustness~\citep{zhang2025two}. Across these techniques, the decorrelation signal acts on hidden features, router weights, parameter residuals, concept embeddings, or latent codes rather than directly on the generative output. Our loss instead operates on the raw noise-prediction outputs of separately parameterized components combined through linear score composition.

\section{Method}
\label{sec:Method}

We introduce a training-time decorrelation mechanism for compositional diffusion policies. The composed prediction is supervised and its decomposition is weakly constrained, allowing multiple factors to learn overlapping denoising directions. We first review the policy setup, then explain the motivation from projective composition, and finally present our practical loss and affine router.

\subsection{Probabilistic Policy Modeling}

Consider offline imitation learning from demonstrations $\mathcal{D}=\{(o_t,a_t^0)\}$, where $o_t$ is the observation and $a_t^0\in\mathbb{R}^{H\times d_a}$ is a clean action chunk. At diffusion step $k$, we construct $a_t^k=\sqrt{\bar{\alpha}_k}a_t^0+\sigma_k\xi$, where $\sigma_k=\sqrt{1-\bar{\alpha}_k}$ and $\xi\sim\mathcal{N}(0,I)$. The denoiser is trained to predict $\xi$ with the standard noise-prediction objective. For variance-preserving diffusion, the corresponding score estimate is $s_\theta^k=-\epsilon_\theta/\sigma_k$, so linear score composition can be implemented through noise predictions.

Following FDP~\citep{liu2026flexible}, the distributional motivation is a weighted product of factors, $\tilde{p}^k(a\mid o_t)\propto \prod_i p_i^k(a\mid o_t)^{w_i(o_t)}$, whose score, when normalizable, is the weighted sum of component scores. Our compositional denoiser has $K$ jointly trained components and router weights $w_i(o_t)$:

\begin{equation}
\label{eq:composed-denoiser}
    \bar{\epsilon}_\theta(a_t^k,o_t,k)
    =\sum_{i=1}^K
    w_i(o_t)\epsilon_{\theta_i}(a_t^k,o_t,k).
\end{equation}

Training supervises the composed prediction, and inference uses it in the usual DDPM reverse update. The components, which we also call factors, are additive predictions supervised only through their weighted sum: they do not need to individually estimate scores of conditional distributions.

\subsection{Factorized Score Composition}

Projective composition~\citep{bradley2025mechanisms} formalizes the goal of preserving components' behavior along specified projections. Under the coordinate-factorized case of Factorized Conditionals, the score deltas $v_i^k=s_i^k-s_b^k$ have disjoint coordinate support and satisfy $\langle v_i^k,v_j^k\rangle=0$ for $i\neq j$. This is a consequence of that sufficient condition, not a general requirement for successful composition. In noise-prediction form, $v_i^k=-(\epsilon_i-\epsilon_b)/\sigma_k$, where $\epsilon_i$ and $\epsilon_b$ are the noise predictions for the component and background scores, so the theoretical orthogonality concerns noise deltas, not raw predictions.

Our architecture does not define an explicit background predictor. Its components provide additive contributions to the composed prediction, without subtracting a background prediction to form residuals. We therefore use raw-output decorrelation as a background-agnostic proxy for specialization. Sum-to-one router weights remove the explicit background correction from the weighted composition formula, but do not make output orthogonality equivalent to delta orthogonality. More on the distinction between component scores and background-relative deltas can be found in Appendix~\ref{app:background-deltas}.

\subsection{Practical Score Decorrelation Loss}

Let $e_i=\mathrm{vec}(\epsilon_{\theta_i}(a_t^k,o_t,k)) \in\mathbb{R}^{Hd_a}$ be the raw prediction of factor $i$ flattened over the full action chunk. Define its normalized envelope as $q_i=|e_i|/(\|e_i\|_2+\delta)$, where $\delta>0$ stabilizes normalization. Our final training objective combines noise prediction with the auxiliary decorrelation loss:
\begin{equation}
\label{eq:training-objective}
    \mathcal{L}
    =\mathbb{E}_{(o_t,a_t^0),k,\xi}
    \left[
    \left\|\xi-\bar{\epsilon}_\theta(a_t^k,o_t,k)\right\|_2^2
    +\frac{2\lambda}{K(K-1)}
    \sum_{i<j}(q_i^\top q_j)^2
    \right].
\end{equation}

For $\tilde{e}_i=e_i/(\|e_i\|_2+\delta)$, the bound $|\tilde{e}_i^\top\tilde{e}_j|\leq q_i^\top q_j$ shows that envelope overlap controls the corresponding signed overlap while preventing cancellation between coordinates. This encourages disjoint support among raw predictions, motivated by coordinate-factorized deltas. Flattening across the full $(H\times d_a)$ chunk makes the penalty act on each component's complete prediction. We convert router logits $\ell_i(o_t)$ into affine weights $w_i(o_t)=\ell_i(o_t)-\frac{1}{K}\sum_j\ell_j(o_t)+\frac{1}{K}$. This enforces $\sum_i w_i(o_t)=1$ while allowing signed weights.

Training follows the standard compositional policy procedure~\citep{liu2026flexible}: sample a demonstration, timestep, and noise, compute component predictions, compose them with router weights, and optimize the combined objective. At inference, the auxiliary loss is discarded and the sampler uses the same composed denoiser. The pairwise penalty adds $O(K^2Hd_a)$ training computation, with no additional sampling cost or required change to the component networks.

\vspace{1pt}
\section{Experiments}
We first study a controlled trajectory-planning task in which the effect of our loss can be visualized directly (Sec.~\ref{sec:toy}). We then apply the loss to Factorized Diffusion Policy (FDP)~\citep{liu2026flexible}, a representative compositional diffusion policy with $K$ factors and a learned router, comparing against unregularized FDP, Diffusion Policy (DP)~\citep{chi2024diffusion} as a non-compositional baseline, and Sparse Diffusion Policy (SDP)~\citep{wang2024sparse} as a mixture-of-experts baseline. We test three claims: decorrelation (1) preserves or improves success on RLBench MT8, RoboCasa atomics, LIBERO-10 composite tasks, and a real robot (Sec.~\ref{sec:main}); (2) responds consistently across factor counts and penalty strengths, and outperforms alternative decorrelation objectives (Sec.~\ref{sec:sensitivity}); and (3) yields factors that specialize to distinct robot action groups (Sec.~\ref{sec:specialization}). $\pm$ denotes the standard error of the mean over evaluation seeds.

\textbf{Metrics.} We measure overlap between factors in three ways. $|\cos|$ is the mean pairwise absolute cosine similarity between factor predictions, near 1 when they are parallel. The double-counting cross-term $D=\frac{2}{H}\sum_{i<j}w_iw_j\langle e_i,e_j\rangle$ (2-count in tables), averaged over evaluation samples and diffusion steps, satisfies $\frac{1}{H}\|\sum_i w_i e_i\|_2^2=\frac{1}{H}\sum_i w_i^2\|e_i\|_2^2+D$, so it is the part of the composed prediction's squared norm that comes from factors reinforcing each other: large $D$ means factors repeat one another, while $D\approx0$ means each contributes on its own. Since signed terms can cancel, a small $D$ can still hide overlap that $|\cos|$ catches. Finally, effective rank, the exponentiated Shannon entropy of the factor covariance eigenvalues, counts how many factors contribute independent directions, from 1 (all copies of each other) to $K$ (all distinct).

\subsection{Controlled Demonstration: Trajectory Planning}
\label{sec:toy}

\begin{figure}[t]
  \centering
  \sbox{\liberotablebox}{%
    \small\renewcommand{\arraystretch}{1.1}%
    \begin{tabular}{lccccc}
      \toprule
      $\lambda$ & 2D & 1D & 3D & $|\cos|$ & 2-count \\
      \midrule
      $0$     & 1.0 & 0.16 & 0.0  & 0.64 & 0.15 \\
      $0.025$ & 1.0 & 0.77 & 0.0  & 0.57 & 0.16 \\
      $0.045$ & 1.0 & 0.94 & 0.16 & 0.47 & 0.13 \\
      $0.1$   & 1.0 & \textbf{0.97} & 0.46 & 0.39 & 0.06 \\
      $1.0$   & 1.0 & 0.90 & \textbf{0.99} & 0.29 & 0.00 \\
      \bottomrule
    \end{tabular}}%
  \setlength{\liberofigwidth}{\dimexpr\textwidth-\wd\liberotablebox-0.75em\relax}%
  \sbox{\liberoplotbox}{\includegraphics[width=\liberofigwidth]{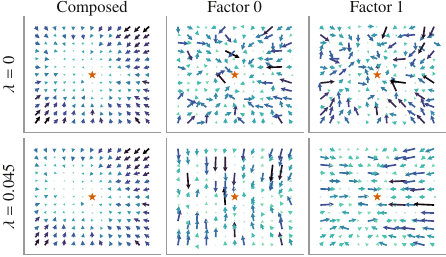}}%
  \begin{minipage}[t]{\liberofigwidth}
    \vspace{0pt}
    \usebox{\liberoplotbox}
    \caption{\textbf{Per-factor score fields for 2D navigation task.} Without our loss (top), both factors redundantly point at the goal. With it (bottom), they split into perpendicular fields without affecting their resulting composition.}
    \label{fig:quiver}
  \end{minipage}\hfill
  \begin{minipage}[t]{\wd\liberotablebox}
    \vspace{2pt}
    \vbox to \ht\liberoplotbox{\vfill\usebox{\liberotablebox}\vfill}
    \captionof{table}{\textbf{Success rates and decorrelation metrics across $\lambda$.} 2D is in-distribution; 1D and 3D are zero-shot (OOD). Higher $\lambda$ lowers overlap and raises zero-shot success.}
    \label{tab:toy}
  \end{minipage}
  \par\vspace{-10pt}
\end{figure}

We first study a controlled 2D trajectory planning task to demonstrate the effects of our loss. A diffusion model must navigate from a uniformly sampled start to a goal in $[-1,1]^2$. The model has two factors, each a four-layer MLP that denoises a 24-step trajectory, combined via linear score composition under a learned router. We train with and without our loss, sweeping $\lambda$.

Figure~\ref{fig:quiver} shows the per-factor fields, with both composed fields clearly pointing towards the goal at $(0,0)$. Without our loss ($\lambda=0$), both factors produce redundant goal-pointing fields with little interpretability. Under decorrelation ($\lambda=0.045$), the factors separate into orthogonal axis-aligned fields (up/down and left/right), recovering the two coordinate axes of the task without axis-level supervision. Table~\ref{tab:toy} shows that cosine similarity and double-counting fall as $\lambda$ increases.

The recovered basis then enables compositional generalization. Running a single factor on pure-horizontal or pure-vertical problems without retraining lifts success from $0.16$ to $0.97$ as $\lambda$ increases. Similarly, applying this 1D primitive along each axis of a 3D task without 3D training lifts success from $0.0$ to $0.99$, peaking exactly as double-counting reaches zero.

Although this toy task could also be solved by composing separately trained 1D primitives, robot action dimensions are coupled and cannot be trained in isolation. Recovering an axis-aligned basis from a jointly trained 2D model suggests our loss can surface such bases in robotic settings too.
  
\subsection{Multitask and Composite Performance}
\label{sec:main}

We evaluate our loss with $K{=}4$ factors on RLBench MT8, the eight-task suite that FDP is originally evaluated on (omitting ``Pick and Place'', on which all models score near 0\%), and on RoboCasa atomics, a seven-task subset of harder robotic tasks. We sweep $\lambda \in \{0.5, 1.25, 2.5\} \times 10^{-3}$ and report success rate alongside the structural diagnostics, computed on a held-out validation split.

\begin{table*}[t]
\centering
\normalsize
\renewcommand{\arraystretch}{1.05}
\setlength{\tabcolsep}{2.5pt}

\resizebox{\textwidth}{!}{%

\begin{tabular}{lccccccccc}
\toprule
& \multicolumn{8}{c}{\textbf{RLBench}} & \\
\cmidrule(lr){2-9}

\textbf{Policy}
& Close Box
& Close Drawer
& Close Fridge
& Close Microwave
& Toilet Seat Down
& Take Umbrella
& Reach Target
& \textbf{Avg.} \\

\midrule

DP
& $9.60_{\pm 1.60}$
& $99.1_{\pm 0.80}$
& $71.2_{\pm 4.12}$
& $51.2_{\pm 2.33}$
& $52.8_{\pm 4.2}$
& $13.6_{\pm 0.90}$
& $16.0_{\pm 2.82}$
& $44.8$ \\

FDP
& 36.8$_{\pm 1.50}$
& 97.6$_{\pm 1.60}$
& 71.2$_{\pm 1.94}$
& \textbf{80.0}$_{\pm 1.79}$
& \textbf{68.0}$_{\pm 2.51}$
& 15.2$_{\pm 2.27}$
& \textbf{24.0}$_{\pm 2.50}$
& \textbf{56.1} \\

FDP + $\lambda_{0.5}$
& \textbf{49.6}$_{\pm  3.25}$
& 97.6$_{\pm 1.60}$
& \textbf{81.6}$_{\pm 3.15}$
& 67.2$_{\pm 4.07}$
& 66.4$_{\pm 4.45}$
& 10.4$_{\pm 3.11}$
& 14.4$_{\pm 3.71}$
& 55.3 \\

FDP + $\lambda_{1.25}$
& 35.2$_{\pm 1.50}$
& \textbf{99.2}$_{\pm 0.80}$
& 73.6$_{\pm 1.92}$
& 75.2$_{\pm 2.37}$
& 56.0$_{\pm 1.53}$
& \textbf{17.6}$_{\pm 1.83}$
& 17.6$_{\pm 1.59}$
& 53.5 \\

FDP + $\lambda_{2.5}$
& 39.2$_{\pm 2.87}$
& 98.4$_{\pm 1.60}$
& 63.2$_{\pm 2.96}$
& 67.2$_{\pm 1.66}$
& 62.4$_{\pm 2.30}$
& 16.8$_{\pm 1.20}$
& 23.2$_{\pm 2.37}$
& 52.9 \\

\bottomrule
\end{tabular}

}

\vspace{0.8em}

\resizebox{\textwidth}{!}{%

\begin{tabular}{lccccccccc}
\toprule
& \multicolumn{8}{c}{\textbf{RoboCasa}} & \\
\cmidrule(lr){2-9}

\textbf{Policy}
& Close Microwave
& Dishwasher Rack
& Close Oven
& Open Microwave
& Open Dishwasher
& On Microwave
& On Stove
& \textbf{Avg.} \\

\midrule

DP
& 32.0$_{\pm 2.20}$
& 24.1$_{\pm 3.10}$
& 25.4$_{\pm 3.16}$
& 8.1$_{\pm 2.00}$
& 8.0$_{\pm 2.54}$
& 4.4$_{\pm 1.87}$
& 4.1$_{\pm 1.55}$
& 15.2 \\

FDP
& 51.0$_{\pm 1.22}$
& 26.2$_{\pm 2.17}$
& 26.0$_{\pm 1.81}$
& 9.1$_{\pm 1.87}$
& 5.0$_{\pm 0.20}$
& 4.0$_{\pm 1.52}$
& 3.2$_{\pm 0.54}$
& 17.8 \\

FDP + $\lambda_{0.5}$
& \textbf{53.2}$_{\pm 1.88}$
& \textbf{30.0}$_{\pm 2.19}$
& \textbf{37.2}$_{\pm 2.89}$
& 15.5$_{\pm 2.74}$
& \textbf{12.8}$_{\pm 2.00}$
& 7.0$_{\pm 0.78}$
& 7.0$_{\pm 1.21}$
& \textbf{23.2} \\

FDP + $\lambda_{1.25}$
& 48.2$_{\pm 1.30}$
& 24.5$_{\pm 2.02}$
& 32.5$_{\pm 2.55}$
& \textbf{24.0}$_{\pm 1.52}$
& 11.7$_{\pm 1.91}$
& 5.2$_{\pm 0.97}$
& 7.0$_{\pm 1.11}$
& 21.9 \\

FDP + $\lambda_{2.5}$
& $48.4_{\pm 2.01}$
& $22.9_{\pm 2.61}$
& $34.0_{\pm 2.30}$
& $19.7_{\pm 1.20}$
& $10.0_{\pm 1.76}$
& $\textbf{9.0}_{\pm 2.20}$
& $\textbf{11.1}_{\pm 1.87}$
& $22.2$ \\

\bottomrule
\end{tabular}

}

\caption{\textbf{Task success rates.} Mean rates across RLBench and RoboCasa atomics, averaged over 10 seeds with 220 episodes each. $\lambda$ values are $\times 10^{-3}$. Decorrelation stays close to FDP on MT8 and improves RoboCasa.}
\label{tab:main_results}
\vspace{-15pt}
\end{table*}

Table~\ref{tab:main_results} reports per-task success rates. On MT8, our best setting comes within a point of the FDP baseline on average (55.3 versus 56.1), and all tested strengths stay within 3.2 points of it while remaining well above the monolithic Diffusion Policy baseline (44.8), indicating that decorrelation does not substantially hurt performance on simpler benchmarks. On RoboCasa, decorrelation produces a more substantial improvement ($17.8\%$ to $23.2\%$), with gains concentrated in the harder tasks (Close Oven, Open Microwave, Open Dishwasher), and higher values of $\lambda$ retain most of this improvement. We attribute the difference in benchmark behavior to dataset scale and task diversity: MT8's $400$ demonstrations across eight short-horizon tasks leave little redundant representation for our loss to compress, while RoboCasa's $5{,}000$ demonstrations across more diverse, kitchen-based atomic tasks benefit more directly from a functionally cleaner factor decomposition.

\begin{table}[H]
  \setlength{\belowcaptionskip}{0pt}
  \begin{minipage}{0.48\textwidth}
    \centering
    \small
    \begin{tabular}{lccc}
      \hline
      \textbf{Model} & $|\cos|$ & 2-count & eff-rank \\
      \hline
      FDP & 0.98 & 9.87 & 1.07 \\
      FDP+$\lambda_{0.5}$ & 0.32 & 4.14 & 3.63 \\
      FDP+$\lambda_{1.25}$ & 0.24 & 2.73 & 3.80 \\
      FDP+$\lambda_{2.5}$ & 0.19 & 1.68 & 3.86 \\
      \hline
    \end{tabular}
    \vspace{4pt}
    \captionof{table}{\textbf{Factor overlap on RLBench MT8.} Our loss reduces overlap on all three metrics.}
    \label{tab:mt8_metrics}
  \end{minipage}\hfill
  \begin{minipage}{0.48\textwidth}
    \centering
    \small
    \begin{tabular}{lccc}
      \hline
      \textbf{Model} & $|\cos|$ & 2-count & eff-rank \\
      \hline
      FDP & 0.97 & 12.2 & 1.12 \\
      FDP+$\lambda_{0.5}$ & 0.25 & 2.67 & 2.57 \\
      FDP+$\lambda_{1.25}$ & 0.21 & 1.48 & 3.21 \\
      FDP+$\lambda_{2.5}$ & 0.04 & 0.17 & 3.14 \\
      \hline
    \end{tabular}
    \vspace{4pt}
    \captionof{table}{\textbf{Factor overlap on RoboCasa atomic.} Factors go from near-equal to near-orthogonal.}
    \label{tab:robocasa_metrics}
  \end{minipage}
\end{table}
\vspace{-\intextsep}

Tables~\ref{tab:mt8_metrics} and~\ref{tab:robocasa_metrics} report the overlap metrics. Without our regularizer, factors collapse to near-identical predictions: pairwise cosine plateaus at $0.97$--$0.98$ and effective rank at roughly $1.1$ on both benchmarks, indicating that all four factors are effectively a single cloned component. Even our smallest tested $\lambda$ dramatically separates them, reducing $|\cos|$ to $0.32$ on MT8 and $0.25$ on RoboCasa, lowering double-counting by more than half, and raising effective rank above $2.5$. As $\lambda$ increases further, these diagnostics improve almost monotonically, with RoboCasa at $\lambda{=}2.5{\times}10^{-3}$ reaching $|\cos|{=}0.04$, near-zero double-counting, and effective rank above $3$. Our loss converts redundant decompositions into structurally clean ones while preserving (and on RoboCasa,
improving) mean success.

\begin{wraptable}{r}{0.55\textwidth}
\vspace{-12pt}
\centering \small
\setlength{\tabcolsep}{4pt} \renewcommand{\arraystretch}{1.05}
\begin{tabular}{lcccc}
\toprule
& \multicolumn{2}{c}{\textbf{Composite}} & \multicolumn{2}{c}{\textbf{Subgoal}}\\
\cmidrule(lr){2-3}\cmidrule(lr){4-5}
$\lambda$ ($\times 10^{-3}$) & $K{=}4$ & $K{=}8$ & $K{=}4$ & $K{=}8$\\
\midrule
0   & $15.3_{\pm2.5}$ & $21.7_{\pm1.7}$ & $24.4_{\pm0.4}$ & $35.3_{\pm1.5}$\\
0.3 & $\mathbf{23.1}_{\pm2.4}$ & $23.9_{\pm1.0}$ & $\mathbf{29.3}_{\pm1.0}$ & $35.3_{\pm1.0}$\\
1   & $15.8_{\pm1.3}$ & $29.2_{\pm0.5}$ & $26.5_{\pm0.7}$ & $37.2_{\pm1.2}$\\
3   & $10.8_{\pm1.7}$ & $\mathbf{35.3}_{\pm2.0}$ & $20.1_{\pm0.7}$ & $\mathbf{46.7}_{\pm2.5}$\\
10  & $16.4_{\pm1.0}$ & $20.3_{\pm0.6}$ & $24.0_{\pm2.4}$ & $35.3_{\pm0.8}$\\
\bottomrule
\end{tabular}
\captionof{table}{\textbf{Composite and subgoal success} on LIBERO-10. Composite requires completing both subgoals, while subgoal gives credit for each: the best $\lambda$ improves both.}
\label{tab:libero_comp}
\vspace{-10pt}
\end{wraptable}

We next test whether decorrelation improves execution of multi-step tasks. We train FDP with $K\in\{4,8\}$ on six LIBERO-10 tasks using 300 demonstrations, 50 per task. Each task combines two subgoals, such as placing two grocery items in a basket or putting a bowl in a drawer and then closing it. Table~\ref{tab:libero_comp} reports two evaluations of the trained checkpoints. Composite success requires completing both subgoals within 650 steps, evaluated over $3\times120$ rollouts. Subgoal success is measured in fresh rollouts in the same scenes, retaining the full-task instruction but scoring with partial credit on both subgoals.

For $K=8$, regularization at $\lambda=3\times10^{-3}$ raises composite success from $21.7\%$ to $35.3\%$ and subgoal success from $35.3\%$ to $46.7\%$. For $K=4$, the best tested strength is lower, $\lambda=3\times10^{-4}$, improving the corresponding rates from $15.3\%$ to $23.1\%$ and from $24.4\%$ to $29.3\%$. Appropriately weighted decorrelation thus improves both complete-task execution and subgoal attainment.

\begin{figure}[t]
  \centering
  \setlength{\tabcolsep}{1pt}
  \renewcommand{\arraystretch}{0}

  \begin{tabular}{@{}cccccc@{}}
    \multicolumn{6}{@{}c@{}}{%
      \begin{tikzpicture}
        \draw[-{Latex[length=7pt, width=6pt]}, line width=1.2pt]
          (0,0) -- ({0.948\textwidth+10pt},0);
      \end{tikzpicture}}
    \\[3pt]
    \includegraphics[width=0.158\textwidth]{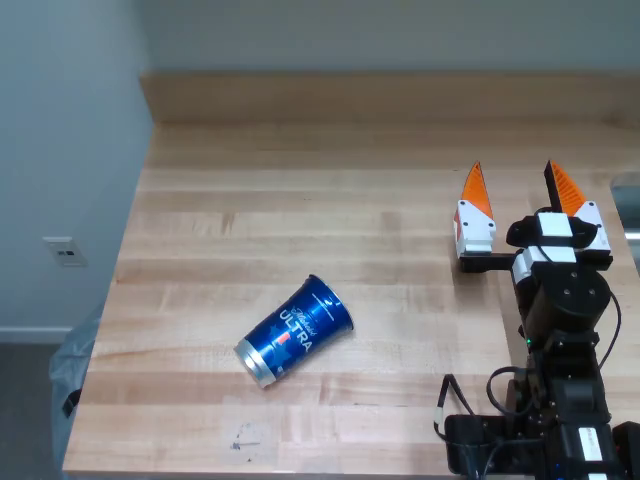} &
    \includegraphics[width=0.158\textwidth]{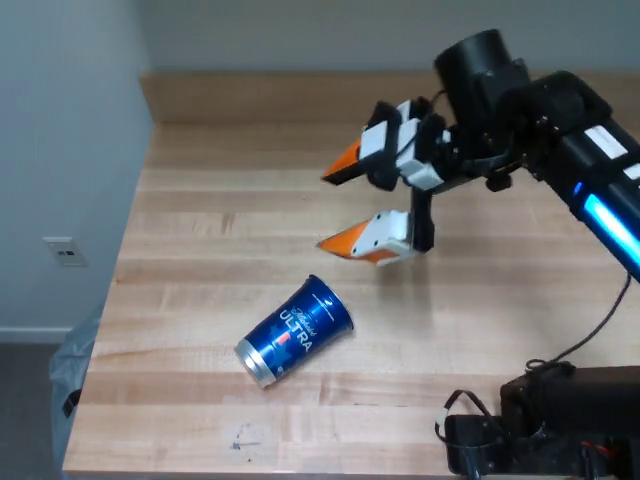} &
    \includegraphics[width=0.158\textwidth]{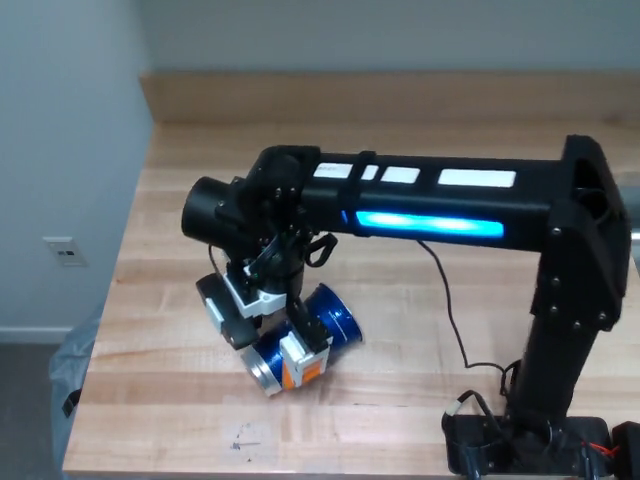} &
    \includegraphics[width=0.158\textwidth]{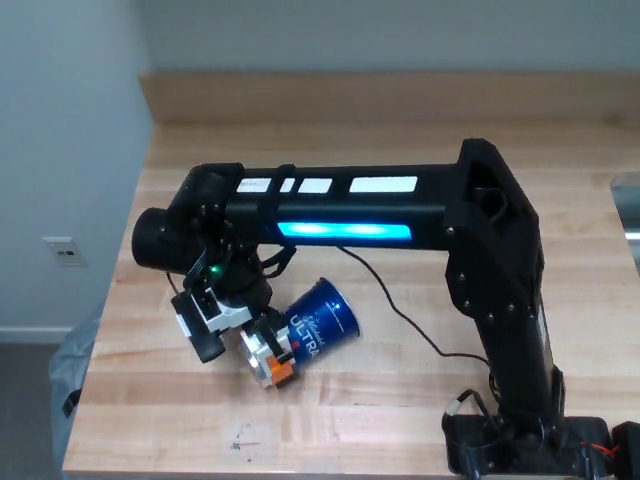} &
    \includegraphics[width=0.158\textwidth]{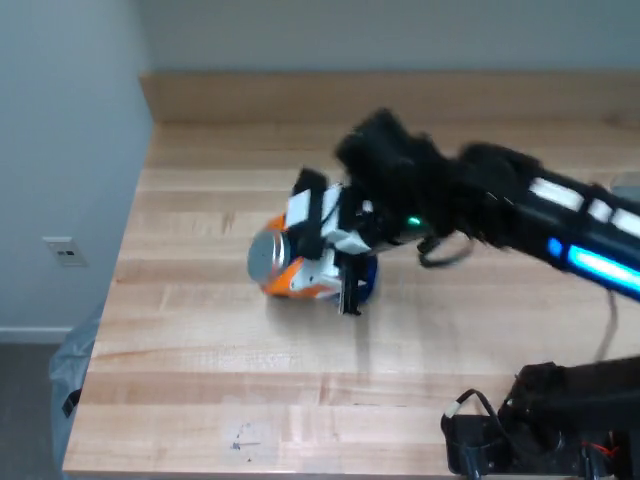} &
    \includegraphics[width=0.158\textwidth]{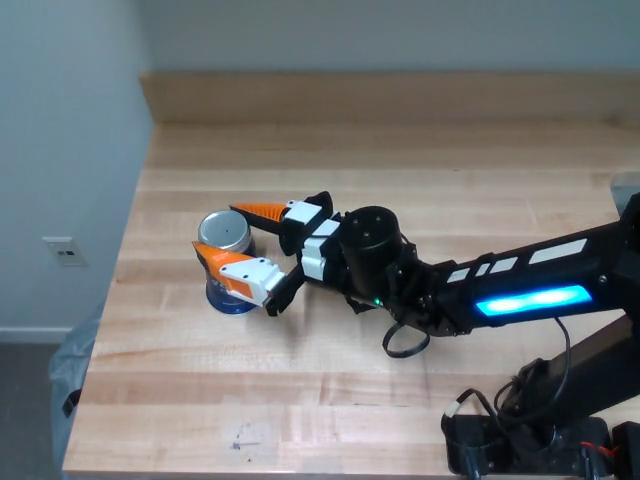}
    \\[2pt]
    \includegraphics[width=0.158\textwidth]{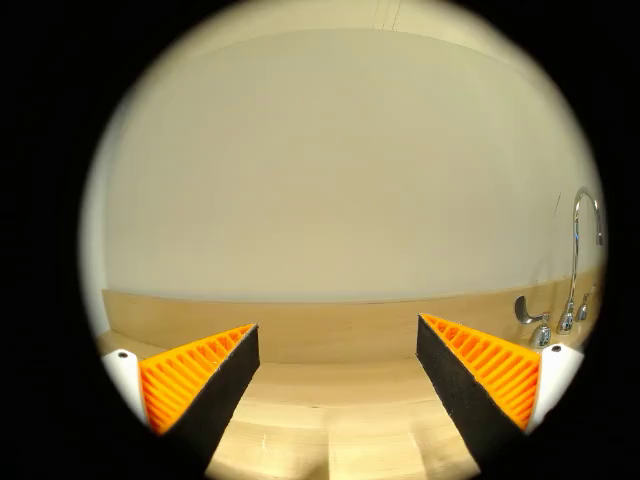} &
    \includegraphics[width=0.158\textwidth]{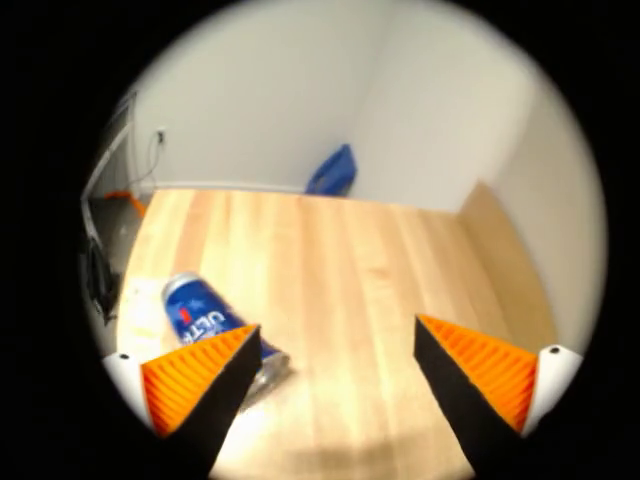} &
    \includegraphics[width=0.158\textwidth]{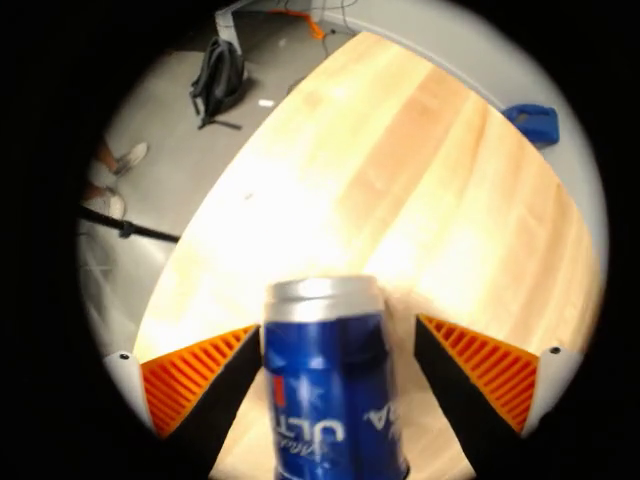} &
    \includegraphics[width=0.158\textwidth]{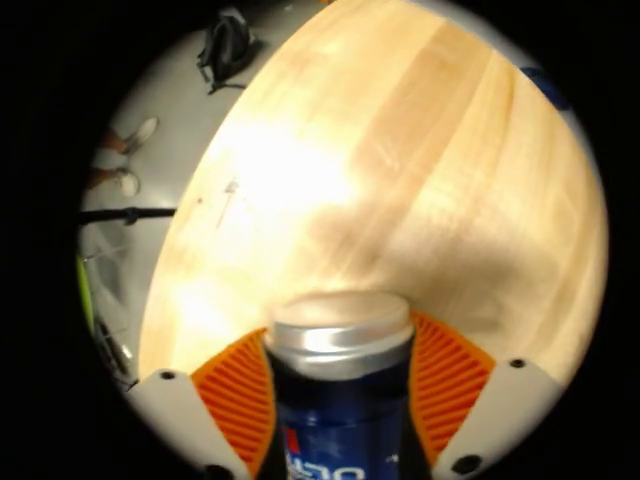} &
    \includegraphics[width=0.158\textwidth]{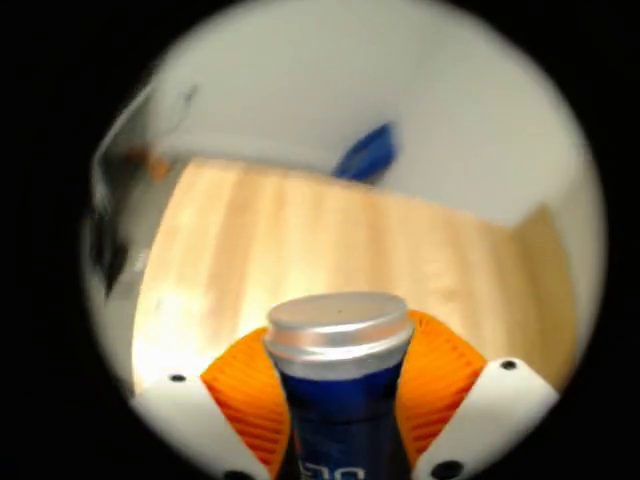} &
    \includegraphics[width=0.158\textwidth]{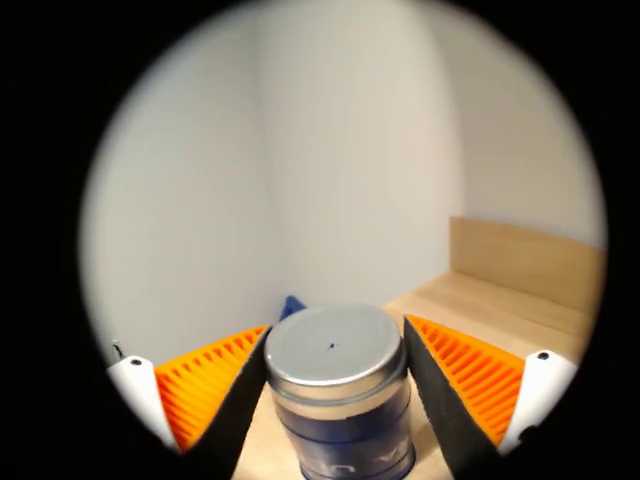}
  \end{tabular}

  \caption{\textbf{Real-world cup flipping.} Agent-view (top) and wrist-view (bottom) frames, left to right, on an ARX R5 arm. Over 25 trials, FDP with our loss succeeds 19 times, versus 15 for FDP and 10 for DP.}
\label{fig:real_robot}
\vspace{-15pt}
\end{figure}

Finally, we verify that these benefits survive real-world dynamics by training a robotic arm to flip a cup using our model. We teleoperate 150 demonstrations of an ARX R5 arm flipping an overturned cup upright (Figure~\ref{fig:real_robot}) and train a DP baseline and two $K{=}8$ FDP policies with $\lambda \in \{0, 10^{-3}\}$ under matched compute, each using an overhead camera and an eye-in-hand camera. Over 25 evaluation rollouts each, DP succeeds in $10/25$, FDP without decorrelation in $15/25$, and FDP with decorrelation in $19/25$. The ordering matches our simulated benchmarks, with decorrelation improving over the undecorrelated model and both FDP variants outperforming the monolithic baseline.

\subsection{Sensitivity to Factor Count, Penalty Strength, and Regularizer Choice}
\label{sec:sensitivity}
\noindent\par 

\begin{figure}[t]
  \centering
  \sbox{\liberotablebox}{%
    \small\setlength{\tabcolsep}{2.5pt}\renewcommand{\arraystretch}{1.1}%
    \begin{tabular}{lccccc}
      \toprule
      $\lambda$ & $K{=}3$ & $K{=}4$ & $K{=}5$ & $K{=}8$ & $K{=}12$\\
      \midrule
      0   & $9.1_{\pm0.7}$ & $15.6_{\pm0.9}$ & $16.7_{\pm0.4}$ & $15.5_{\pm0.3}$ & $16.4_{\pm0.6}$\\
      0.3 & $8.4_{\pm0.8}$ & $14.7_{\pm0.9}$ & $\mathbf{18.7}_{\pm0.6}$ & $18.0_{\pm0.9}$ & $\mathbf{19.3}_{\pm0.5}$\\
      1   & $6.7_{\pm0.2}$ & $\mathbf{16.2}_{\pm0.8}$ & $12.5_{\pm1.1}$ & $18.0_{\pm0.6}$ & $17.8_{\pm0.8}$\\
      3   & $\mathbf{10.9}_{\pm1.0}$ & $11.9_{\pm0.8}$ & $12.3_{\pm0.4}$ & $\mathbf{19.0}_{\pm0.1}$ & $17.9_{\pm0.4}$\\
      10  & $5.8_{\pm0.7}$ & $4.7_{\pm0.2}$ & $9.3_{\pm0.3}$ & $8.7_{\pm0.4}$ & $16.8_{\pm0.7}$\\
      \bottomrule
    \end{tabular}}%
  \setlength{\liberofigwidth}{\dimexpr\textwidth-\wd\liberotablebox-1.5em\relax}%
  \sbox{\liberoplotbox}{\includegraphics[width=\liberofigwidth]{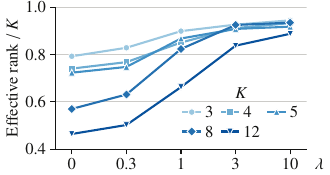}}%
  \begin{minipage}[t]{\wd\liberotablebox}
    \vspace{1pt}
   \vbox to \ht\liberoplotbox{\vskip 11pt\usebox{\liberotablebox}\vfill}
    \captionof{table}{\textbf{Success (\%) on 25 LIBERO-90 atomic tasks across factor counts $K$.} $\lambda$ is $\times 10^{-3}$. The best $\lambda$ in 0.3--3 beats $\lambda{=}0$ at every $K$, most at $K{=}8$ and 12. DP and SDP reach 16.4\% and 16.7\% respectively.}
    \label{tab:libero_k}
  \end{minipage}\hfill
  \begin{minipage}[t]{\liberofigwidth}
    \vspace{0pt}
    \usebox{\liberoplotbox}
    \caption{\textbf{Normalized effective rank on LIBERO-90.} Number of independent directions in the factors' predictions divided by $K$ (factor count). Decorrelation raises it at every $K$, most for large $K$. $\lambda$ is $\times 10^{-3}$.}
    \label{fig:k_sweep_diag}
  \end{minipage}
  \par\vspace{-10pt}
\end{figure}
All experiments in this section use 25 LIBERO-90 atomic tasks with 50 demonstrations per task, evaluated over $3\times500$ rollouts. We first train FDP with $K\in\{3,4,5,8,12\}$ factors across a sweep of $\lambda$. At every factor count, increasing $\lambda$ monotonically increases effective rank$/K$ (Figure~\ref{fig:k_sweep_diag}). Without regularization, effective rank grows more slowly than $K$, so additional factors add diminishing output diversity. Decorrelation mitigates this effect, bringing effective rank above $0.83K$ at every tested factor count for $\lambda=3\times10^{-3}$, up from $0.47K$ at $K{=}12$.

Table~\ref{tab:libero_k} shows that these structural gains can translate into better task performance. The best tested regularized setting improves mean success at every $K$, including $15.5\%\rightarrow19.0\%$ at $K=8$ and $16.4\%\rightarrow19.3\%$ at $K=12$. However, the strongest penalty reduces success below the unregularized baseline for four of the five factor counts, despite continued improvements in the structural diagnostics. This rise and fall, also seen on LIBERO-10 (Table~\ref{tab:libero_comp}), reflects two effects of the penalty. At moderate strength, it removes redundant copies, so factors cover complementary directions. Robot action dimensions are coupled, however, so accurate denoising sometimes needs several factors to contribute to the same coordinates; strong penalties suppress this useful overlap too and compete with the denoising objective. The best $\lambda$ therefore depends on $K$, and output diversity alone does not guarantee better performance.

\begin{figure}[H]
  \centering
  \sbox{\liberotablebox}{%
    \setlength{\tabcolsep}{3.5pt}%
    \begin{tabular}{llcccc}
      \toprule
      \textbf{Objective} & \textbf{Locus} & \textbf{Coef.} & \textbf{SR (\%)} & \textbf{2-count} & \textbf{eff-rank} \\
      \midrule
      None      & ---    & ---                & $15.5_{\pm0.3}$ & 4.74 & 4.56 \\
      RouterVar & router & $10^{-1}$          & $16.9_{\pm0.8}$ & 4.71 & 3.77 \\
      DeCov     & activ. & $10^{-1}$          & $17.9_{\pm0.8}$ & 4.93 & 4.27 \\
      SRIP      & param. & $10^{-2}$          & $17.1_{\pm0.1}$ & 4.77 & 4.54 \\
      NCL       & score  & $10^{-3}$          & $16.5_{\pm0.9}$ & \textbf{1.11} & 3.27 \\
      DeCov     & score  & $10^{-3}$          & $18.0_{\pm0.5}$ & 4.48 & 4.91 \\
      Ours      & score  & $3{\times}10^{-3}$ & $\mathbf{19.0}_{\pm0.1}$ & 2.20 & \textbf{7.40} \\
      \bottomrule
    \end{tabular}}%
  \setlength{\liberofigwidth}{\dimexpr\textwidth-\wd\liberotablebox-1.5em\relax}%
  \sbox{\liberoplotbox}{\includegraphics[width=\liberofigwidth]{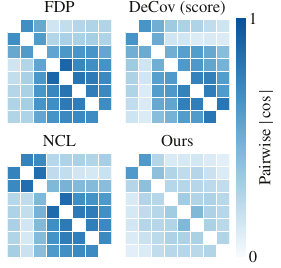}}%
  \begin{minipage}[t]{\wd\liberotablebox}
    \vspace{0pt}
\vbox to \ht\liberoplotbox{\vfill\usebox{\liberotablebox}\vfill}
    \captionof{table}{\textbf{Decorrelation objectives at $K{=}8$ on LIBERO-90.} SR over 3 seeds $\times$ 500 rollouts. Ours gives the highest success and effective rank, keeping the factors distinct.}
    \label{tab:decorr_baselines}
  \end{minipage}\hfill
  \begin{minipage}[t]{\liberofigwidth}
    \vspace{0pt}
    \usebox{\liberoplotbox}
    \caption{\textbf{Pairwise similarity between factors.} Dark blocks mark redundant factors; only our loss removes them.}
    \label{fig:pairwisecos}
  \end{minipage}
  \par\vspace{-15pt}
\end{figure}

We next compare our loss against alternative decorrelation objectives at $K{=}8$, using the best tested coefficient for each baseline:   router-variance regularization (RouterVar)~\citep{guo2026advancing}, the DeCov covariance penalty on activations or score outputs~\citep{cogswell2015reducing}, spectral restricted isometry property regularization (SRIP) on parameters~\citep{bansal2018srip}, and negative correlation learning (NCL) on score outputs~\citep{ncl}. Table~\ref{tab:decorr_baselines} shows that all baselines improve mean success over unregularized FDP, but ours achieves the highest mean ($19.0\%$ versus $16.5$--$18.0\%$) and highest effective rank. Regularization outside the score outputs does not raise effective rank above the FDP baseline ($3.77$--$4.54$ versus $4.56$). Figure~\ref{fig:pairwisecos} shows the overlap between factor predictions directly: without regularization, five of the eight factors form a block of strongly overlapping predictions, which DeCov on score outputs only lightens. NCL achieves the lowest double-counting ($1.11$) but leaves this block largely intact, and its effective rank falls below the FDP baseline ($3.27$ versus $4.56$), showing that reducing this cross-term alone does not increase factor diversity. Our envelope penalty instead removes these blocks, giving an effective rank of $7.40/8$ and supporting its use for encouraging distinct predictions.

\subsection{Behaviorally Meaningful Component Specialization}
\label{sec:specialization}

The previous results show that decorrelation reduces redundancy and improves compositional generalization, but they leave open whether the resulting factor structure is semantically meaningful or merely statistically orthogonal. We address this question on the atomic RoboCasa policy by comparing the same FDP architecture trained with and without our loss ($\lambda=0$ versus $\lambda=5{\times}10^{-4}$), and inspecting how each factor allocates its predicted noise across the 12-dimensional action space.

\begin{figure}
\centering
\includegraphics[width=0.95\linewidth]{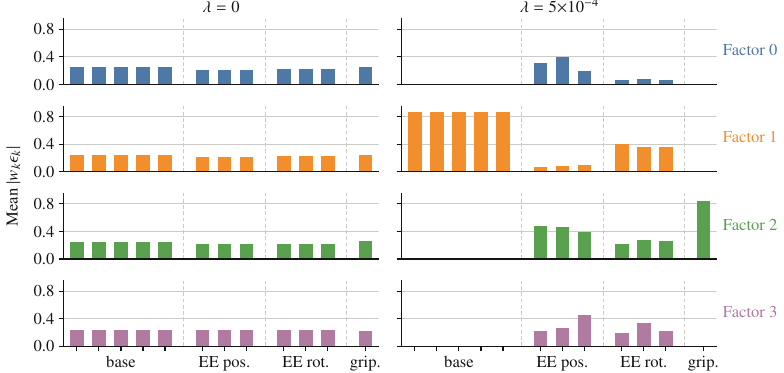}
\caption{\textbf{Per-factor magnitude across action dimensions.} Mean weighted absolute noise prediction of each factor on each action dimension. Without our loss (left), the four factors look identical; with it (right), factor~1 takes the base and factor~2 the gripper, with factors~0 and~3 sharing the arm.}
\label{fig:bars}
    \vspace{-15pt}
\end{figure}

\begin{wrapfigure}{r}{0.35\textwidth}
\vspace{-1\intextsep}
\centering
\includegraphics[width=\linewidth]{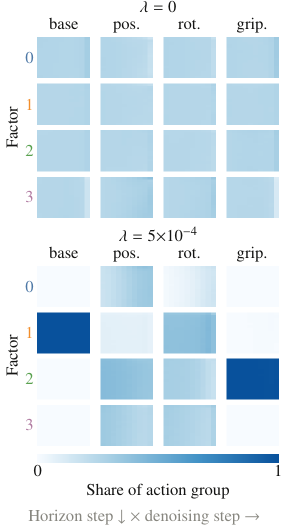}
\caption{\textbf{Action attribution across horizon and denoising steps.} With our loss, factor~1 owns the base and factor~2 the gripper at every step.}
\label{fig:heatmap}
\vspace{-0.5\intextsep}
\end{wrapfigure}

Figure~\ref{fig:bars} shows the mean absolute noise prediction of each factor across action dimensions. Without our loss, the four factors emit nearly identical bar heights on every dimension, where only the router weights distinguish their outputs: functionally, every factor performs every role. Under decorrelation, the router weights collapse to a uniform $0.25$ across factors, yet the factors themselves clearly specialize. Factor~1 carries nearly all of the magnitude on the five base dimensions and factor~2 nearly all of the gripper prediction, while end-effector position and rotation are spread across the factors. The model thus separates base and gripper control from arm control without any supervision over the action space.

Figure~\ref{fig:heatmap} provides a more detailed view of this assignment across the action horizon and denoising trajectory. Each panel corresponds to a factor--action-group pair, with axes indexing horizon position and denoising step, and color indicating the factor's share of contribution to that action group. Without decorrelation, every panel remains near the uniform share of $0.25$. Under decorrelation, factor~1 dominates the base dimensions and factor~2 dominates the gripper, leaving little contribution from other factors to these groups. The consistent color within each panel shows that this dominance persists across horizon positions and noise levels. The factor identities match the bar plot, but the heatmaps additionally show that specialization is not created by averaging over changing assignments. Instead, the same components maintain distinct contributions across action groups throughout the action horizon and plotted denoising trajectory.

\begin{figure}[t]
\centering
\includegraphics[width=0.99\linewidth]{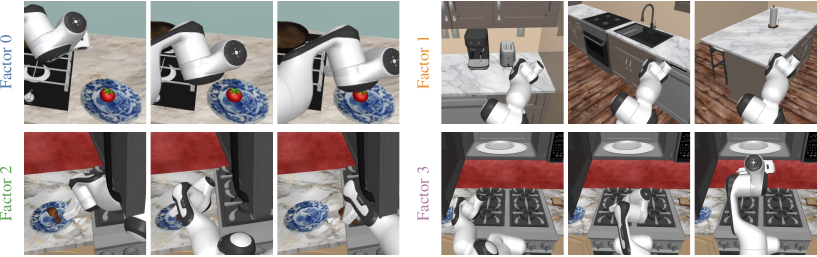}
\caption{\textbf{Rollout snippets of each factor's action channel}, $K{=}4$ with our loss on RoboCasa. Factor 0 makes fine arm moves, 1 drives the base, 2 grasps, and 3 assists large arm moves; time runs left to right.}
\label{fig:factor_strips}
\vspace{-10pt}
\end{figure}

To demonstrate the qualitative effects of each factor's contributions, we separately train the same $K{=}4$ architecture with our loss ($\lambda{=}5{\times}10^{-4}$) on three RoboCasa pick-and-place tasks. Figure~\ref{fig:factor_strips} shows rollout snippets where each factor's channel is in use, showing the same decomposition: factor~2 accounts for 99\% of the gripper prediction while the robot grasps and lifts an object, factor~1 for most of the base command (64\%) while the robot drives across the kitchen, and factor~0 for most of the arm prediction (56\%) during fine adjustments. Factor~3 owns no channel of its own, with its share of arm position growing with arm speed, from 24\% during slow motion to 36\% during the large move shown.

\section{Limitations and Future Work}

Our approach has limitations that point to promising directions for future work. First, the penalty strength is a hyperparameter and must be tuned. The best $\lambda$ varies with the number of factors and the benchmark, and the strongest penalty we test lowers LIBERO-90 success below the unregularized baseline at four of five factor counts even as the structural diagnostics keep improving (Table~\ref{tab:libero_k}). This could be addressed by annealing $\lambda$ during training, scaling it with the noise level, or adapting it online to reach a target effective rank rather than fixing it in advance.

Second, our loss encourages specialization along action coordinates rather than task-level skills. Because it compares elementwise magnitudes, it penalizes contributions to the same action--time coordinates even when they serve different behavioral roles, and on RoboCasa the router weights become uniform once the factors separate, so specialization no longer depends on the observation. Weak subtask annotations could supervise routing and component denoising on the corresponding demonstration segments, applying decorrelation only to jointly active components. Factors would then keep consistent behavioral roles across tasks and could reuse action dimensions without penalizing inactive factors.

Lastly, our loss only approximates its theoretical motivation. Because components are trained jointly through their weighted sum, reduced output overlap does not guarantee that they can be reused in unseen combinations, and raw-output decorrelation is a proxy for the score-delta orthogonality implied by Factorized Conditionals. Learning an explicit background predictor and regularizing residual contributions would align the two more directly, although projective-composition guarantees would still require additional distributional assumptions.

\section{Conclusion}
We present a training-time decorrelation loss that encourages specialization among the additive outputs of compositional diffusion policies. Across RLBench, RoboCasa, and LIBERO, the loss reduces output overlap and promotes more diverse component predictions across factor counts. At suitable regularization strengths, these changes accompany improved atomic and composite task success, higher subgoal completion, and specialization aligned with robot action groups. Comparisons with alternative regularizers support the value of penalizing overlap directly at the component outputs, while a cup-flipping experiment provides initial real-world evidence. The results also show that structural separation alone does not ensure better performance, as useful specialization requires balancing decorrelation with the denoising objective.

\acknowledgments{This work has been made possible in part by a gift from the Chan Zuckerberg Initiative Foundation to establish the Kempner Institute for the Study of Natural and Artificial Intelligence at Harvard University. This work was also supported in part by Meta Robotics Studio, Amazon Robotics, and the Toyota Research Institute (TRI).}

\bibliography{example}  

\clearpage
\appendix

\section*{Appendix}
\section{Background-Relative Deltas and the Practical Loss}
\label{app:background-deltas}

At a fixed observation and noise level, write
$v_i^k=s_i^k-s_b^k$. The weighted composition score is
\begin{align}
    s_C^k
    &=s_b^k+\sum_i w_i v_i^k\\
    &=\sum_i w_i s_i^k+\left(1-\sum_i w_i\right)s_b^k
\end{align}
When $\sum_i w_i=1$, the explicit background correction in the second expression vanishes. However, the deltas in the first expression still depend on the background. In particular,
\begin{align}
    \langle s_i^k,s_j^k\rangle
    &=
    \langle v_i^k,v_j^k\rangle
    +\langle v_i^k,s_b^k\rangle\\
    &\quad+\langle s_b^k,v_j^k\rangle
    +\|s_b^k\|_2^2
\end{align}
Consequently, raw-score orthogonality is not equivalent to delta orthogonality, regardless of the sum-to-one constraint. For variance-preserving diffusion, the corresponding exact noise predictions satisfy $\epsilon_i=-\sigma_k s_i^k$ and $\epsilon_b=-\sigma_k s_b^k$. Therefore,
\begin{equation}
    \langle v_i^k,v_j^k\rangle
    =
    \frac{
    \langle\epsilon_i-\epsilon_b,\epsilon_j-\epsilon_b\rangle
    }{\sigma_k^2}
\end{equation}
An explicit residual-head parameterization would represent
$\rho_i=\epsilon_i-\epsilon_b$ and compose predictions as
\begin{equation}
    \epsilon_C
    =\epsilon_b+\sum_i w_i\rho_i.
\end{equation}
A penalty on $\rho_i$ would directly regularize background-relative noise deltas. Our implementation instead penalizes the raw outputs $\epsilon_{\theta_i}$ of jointly trained additive heads, without constructing $\epsilon_b$ or $\rho_i$. The residual-head interpretation motivates specialization of these additive contributions, but does not establish their equivalence to theoretical deltas.

Even exact delta orthogonality would not by itself establish Factorized Conditionals, which additionally requires distributional independence and a consistent factor structure~\citep{bradley2025mechanisms}. Likewise, the weighted composition identity does not extend the projective-composition guarantee to arbitrary learned weights. Signed weights are admissible in the score algebra, but a density interpretation requires a finite normalizing constant. Our claims should therefore be read as those of an empirically backed background-agnostic regularizer, rather than a guarantee of projective composition.

\section{Loss and Router Design Choices}

We selected our flattened envelope squared-cosine loss and affine router by varying the granularity of the penalty, its treatment of sign, its shape, and the router parameterization. Below we explain the representational and optimization effects of these choices and how they help interpret the observed performance differences. All variants regularize raw component predictions; their connection to background-relative score deltas is motivational, as discussed in Appendix~\ref{app:background-deltas}.

\textbf{Per-step squared cosine.} This variant penalizes squared cosine similarity between component predictions separately at each horizon step. With $K=4$, $d_a=12$, and $H=16$ on RoboCasa, it encourages four mutually orthogonal directions in a 12-dimensional action space at each of 16 steps. This imposes separation locally, even when useful specialization may emerge across different phases of the trajectory: two factors can make similar contributions at one step while playing different roles over the complete action chunk. The local penalty discourages such overlap regardless of their behavior elsewhere. Flattening instead lets differences across the whole trajectory contribute to separation. Empirically, the per-step variant reaches $46.2\%$ success on MT8 and underperforms our chosen loss, consistent with an overly restrictive local separation bias.

\textbf{Per-step squared cosine with softmax router.} This variant uses the same local penalty with nonnegative, sum-to-one router weights. Softmax prevents the router from reversing a learned component's contribution through a negative coefficient. Signed head outputs still permit residual corrections, but the heads must supply the appropriate directions themselves, rather than allowing the router to add or subtract a given output according to the observation. This restricts the ways component predictions can be recombined. Performance is similar to the preceding per-step variant, suggesting that the local orthogonality penalty is the larger limitation in this comparison. 

\textbf{Flat absolute cosine.} This variant moves the penalty to the full $H\cdot d_a$ prediction and penalizes the absolute signed cosine. It avoids requiring separation at every individual step, but retains two limitations. First, the absolute-value penalty has a nonvanishing one-sided slope near zero cosine: it continues to exert pressure when predictions are already nearly orthogonal, potentially competing with the denoising objective and making optimization near zero less stable. Second, positive and negative coordinate products can cancel within the inner product. Two dense predictions can therefore achieve a small penalty while sharing the same trajectory coordinates. These properties offer an explanation for why moving to the full trajectory alone did not match the performance of our envelope-based loss.

\textbf{Flat envelope absolute cosine.}
Applying elementwise absolute values before computing cosine similarity removes cancellation between coordinate products. A low penalty now favors limited co-occupancy of $(h,d)$ cells, rather than allowing dense overlapping predictions whose inner product happens to vanish. This more directly encourages distinct coordinate contributions and improves on the signed alternatives in our comparisons. However, penalizing envelope overlap linearly maintains pressure even when overlap is already small. In coupled robotic action spaces, further suppressing those remaining shared contributions may interfere with accurate denoising. Our chosen squared form softens this pressure as overlap approaches zero, providing a more gradual tradeoff between separation and task-relevant prediction.

\textbf{Flat squared cosine (signed) summed over pairs.} This variant sums squared signed cosine similarities over the $\binom{K}{2}$ pairs of flattened predictions. Squaring reduces the derivative with respect to cosine near zero, addressing the persistent pressure of the absolute-cosine variant. It does not, however, address cancellation between coordinates: the cosine is squared only after positive and negative coordinate products have been summed. Consequently, the penalty cannot distinguish disjoint contributions from overlapping predictions that are orthogonal through cancellation. Its weaker performance than the envelope variant suggests that the treatment of coordinate overlap, not only the penalty shape, matters.

\textbf{Flat ReLU squared cosine summed over pairs.} This variant penalizes only positive cosine similarities, allowing anti-correlated predictions in case they support useful additive and subtractive roles. The resulting zero-penalty region also admits strongly anti-aligned predictions that occupy the same coordinates and encode the same direction up to sign. Such components can contribute through cancellation or signed routing without developing distinct roles. Thus, the objective can be satisfied without reducing the underlying overlap between component outputs. Its lower success in our comparisons supports penalizing shared coordinate occupancy regardless of sign.

Our chosen flattened envelope squared-cosine penalty combines these considerations: it evaluates complete action chunks, prevents cancellation between coordinate products, and reduces regularization pressure as envelope overlap becomes small. The affine router additionally permits signed recombination of the learned outputs. These choices encourage specialization while retaining flexibility for the primary denoising objective.
\end{document}